\documentclass[a4paper,twoside]{article}

\usepackage{calc}
\usepackage{amssymb}
\usepackage{amstext}
\usepackage{amsmath}
\usepackage{amsthm}
\usepackage{multicol}
\usepackage{pslatex}
\usepackage{apalike}
\usepackage{hyperref}
\usepackage{graphicx}
\usepackage{balance}
\usepackage{draftcopy}
\usepackage{color}
\usepackage{SCITEPRESS}  

\begin{document}
\title{\textcolor{black} {Incremental} Principal Component Analysis
\subtitle{Exact implementation and continuity corrections}}

\author{\authorname{Vittorio Lippi\sup{1} and Giacomo Ceccarelli\sup{2}}
\affiliation{\sup{1} \textcolor{black} {Fachgebiet Regelungssysteme Sekretariat EN11, Technische Universit{\"a}t Berlin, 
Einsteinufer 17, Berlin, Germany}}
\affiliation{\sup{2}Dipartimento di Fisica, Universit\`a di Pisa, Largo Bruno Pontecorvo 2, I-56127 Pisa, Italy}
\email{vittorio.lippi@\textcolor{black}{tu-berlin.de}, giacomo.ceccarelli@df.unipi.it}
}


\abstract{This paper describes some applications of an incremental
implementation of the principal component analysis (PCA).
The algorithm updates the transformation coefficients matrix on-line for each
new sample, without the need to keep all the samples in memory.
The algorithm is formally equivalent to the usual batch version, in the sense
that given a sample set the transformation coefficients at the end of the
process are the same.
The implications of applying the PCA in real time are discussed with the
help of data analysis examples.
In particular we focus on the problem of the continuity of the PCs during an
on-line analysis.}

\onecolumn \maketitle \normalsize \vfill

\section{\uppercase{Introduction}}
\label{sec:introduction}
\subsection{Incremental PCA}
Principal Component Analysis (PCA) is a widely used technique and a well-studied subject in the literature. PCA is a technique to reduce data dimensionality of a set of correlated variables. Several natural phenomena and industrial processes are described by a large number of variables and hence their study can benefit from the dimensionality reduction PCA has been invented for. As such PCA naturally applies to statistical data analysis. This means that such technique is traditionally implemented as an offline batch operation. Nevertheless, PCA can be useful when applied to data that are available incrementally, e.g. in the context of process monitoring \cite{dunia1996identification} or gesture recognition \cite{Lippi2009}. 
The PCA can be applied to a data-flow after defining the transformation on a
representative off-line training set set using the batch algorithm \cite{lippi2011can}. This approach can be used in pattern recognition problems for data pre-processing \cite{Lippi2009}. 
Nevertheless one can imagine an on-line implementation of the algorithm.
An on-line implementation is more efficient in terms of memory usage than a batch one. This can be particularly relevant for memory consuming data-sets such as image collections; in fact in the field of visual processing some techniques to implement incremental PCA have been proposed, see for example \cite{artavc2002incremental}.
PCA consists of a linear transformation to be applied to the data-set. Dimensionality reduction is performed by selecting a subset of the transformed variables that are considered more relevant in the sense that they exhibit a larger variance compared to the others. Usually the transformation is calculated and computed on the Z-score, and hence the averages and the variances of the dataset are taken into account. Depending on the applications the algorithm has been extended in different ways, adding samples on-line as presented in \cite{artavc2002incremental} or  incrementally increasing the dimension of the reduced variable subset as seen in \cite{neto2005incremental}. A technique to dynamically merge and split the variable subsets has been presented in \cite{hall2000merging}
Several \textit{approximate} incremental algorithms have been proposed for PCA, e.g. see \cite{shamir2015convergence} and
\cite{boutsidis2015online}, as well as for singular value decomposition \cite{sarwar2002incremental}. An implementation for on-line PCA  has been proposed, for example, for the R language \cite{DegrasandCardot2015}.  In some cases the incremental process is designed to preserve some specific information; for example in \cite{hall1998incremental} the average of the samples is updated with new observations.

Currently, to the best of our knowledge, there is no available description of an exact incremental implementation of PCA, where \textit{exact} means that the transformation obtained given $n$ samples is exactly the same as would have been produced by the batch algorithm, including the z-score normalization, a step that is not included in previous works presenting a similar approach like\cite{artavc2002incremental}.  We decided in light of this to describe the algorithm in detail in this paper. The incremental techniques cited above \cite{hall1998incremental,artavc2002incremental,hall2000merging} are designed to update the reduced set of variables and change its dimensionality when it is convenient for data representation.  In the present work, no indication is provided for which subset of variables should be used , i.e. how many principal components to consider. All the components are used during the algorithm to ensure the exact solution. After describing the exact algorithm some implication of using this incremental analysis are discussed. In particular, we provide an intuitive definition of continuity for the obtained transformation and then we propose a modified version designed  to avoid discontinuities. 

The concept of continuity is strictly related to the incremental nature of the proposed algorithm: in standard PCA the batch analysis implies that the notion of time does not exist, e.g. the order of the elements in the sample set is not relevant for the batch algorithm. In our treatment we instead want to follow the time evolution of variances and eigenvectors.
We are thus lead to consider a dynamical evolution. 

The paper is organized as follows.
In the remaining part of the Introduction we recall the PCA algorithm and we introduce the notation used.
In Section \ref{sec:incremental} we give a detailed account  of the incremental
algorithm for an on-line use of PCA. 
In Section \ref{sec:continuity} we address the problems related to the data
reconstruction, in particular those connected with the signal continuity.
In Sections \ref{sec:results}, \ref{sec:conclusion} we then present the results
of some applications to an industrial data set and draw our conclusions.

\subsection{The principal component analysis}
\label{sec:PCA}
The computation for the PCA starts considering a set of observed data.
We suppose we have $m$ sensors which sample  some physical observables at constant rate.
After $n$ observations we can construct the matrix
\begin{equation}
X_n=\left[\begin{array}{c}
            x_1    \\
            x_2    \\
            \vdots \\
            x_n
          \end{array}\right]
\end{equation}
where $x_i$ is a row vector of length $m$ representing the measurements of the
$i^{th}$ time step so that $X_n$ is a $n \times m$ real matrix whose columns
represent all the values of a given observable.

The next step is to define the sample means $\bar{x}_n$ and standard deviations
$\sigma_n$ with respect to the columns (i.e. for the observables) in the usual
way as
\begin{align}
\bar{x}_{n(j)} &= \frac{1}{n} \sum^n_{i=1} X_{n(ij)} \\
\sigma_{n(j)}  &= \sqrt{\frac{1}{n-1} \sum^{n}_{i=1}
                  \left[ X_{n(ij)} - \bar{x}_{n(j)} \right]^2}
\end{align}
where in parentheses we write the matrix and vector indices explicitly.
In this way we can define the standardized matrix for the data as
\begin{equation}
Z_n=\left[\begin{array}{c}
            x_1-\bar{x}_n \\
            x_2-\bar{x}_n \\
            \vdots        \\
            x_n-\bar{x}_n
          \end{array}\right]\Sigma_n^{-1}
\end{equation}
where $\Sigma_n \equiv {\rm diag}(\sigma_n)$ is a $m \times m$ matrix.
The covariance matrix $Q_n$ of the data matrix $X_n$ is then defined as
\begin{equation}
\label{eq:cov}
Q^{}_n = \frac{1}{n-1} Z^{T}_n Z^{}_n \; .
\end{equation}
We see that $Q_n$ is for any $n$ a symmetric $m \times m$ matrix and it is
positive definite.

Finally we make a standard diagonalization so that we can write
\begin{equation}
Q_n = C^{-1}_n \left[\begin{array}{cccc}
                  \lambda_1 \\
              &   \lambda_2 \\
              &&  \ddots \\
              &&& \lambda_m
                  \end{array}\right] C_n
\end{equation}
where the (positive) eigenvalues $\lambda_i$ are in descending order:
$\lambda_i > \lambda_{i+1}$.
The transformation matrix $C_n$ is the eigenvectors matrix and it is
orthogonal, $C^{-1}_n = C^T_n$.
Its rows are the principal components of the matrix $Q_n$ and the value of 
$\lambda_i$ represents the variance associated to the $i^{th}$ principal
component.
Setting $P_n = Z_n C_n$, we have a time evolution for the values of the PCs
until time step $n$.

We recall that the diagonalization procedure is not uniquely defined: once the
order of the eigenvalues is chosen, one can still choose the ``sign'' of the
eigenvector for one-dimensional eigenspaces and a suitable orthonormal basis for
degenerate ones (in Section \ref{sec:continuity} we will see some consequences
of this fact).
We stress that, since only the eigenspace structure is an intrinsic property
of the data, the PCs are quantity useful for their interpretation but they are
not uniquely defined.

\section{On-line analysis}
\subsection{Incremental algorithm}
\label{sec:incremental}

The aim of the algorithm is to construct the covariance matrix $Q_{n+1}$
starting from the old matrix $Q_n$ and the new observed data $x_{n+1}$.
To do this, at the beginning of step $(n+1)$, we consider the sums of the
observables and their squares after step $n$:

\begin{equation}
a_{n(j)} = \sum^{n}_{i=1} X_{n(ij)}
\end{equation}

\begin{equation}
b_{n(j)} = \sum^{n}_{i=1} X_{n(ij)}^2
\end{equation}

These sums are updated on-line at every step.
From these quantities we can recover the starting means and standard
deviations: $\bar{x}_n = a_n / n$ and $(n-1) \sigma^2_n = b_n - n a^2_n $.
Similarly the current means and standard deviations are also simply obtained.

The key observation to get an incremental algorithm is the following identity:
\begin{equation}
Z_{n+1} = \left[ \begin{array}{c}
                 Z_n \Sigma_n + \Delta \\
                 y
                 \end{array} \right] \Sigma^{-1}_{n+1}
\end{equation}
where $y = x_{n+1} - \bar{x}_{n+1}$ is a row vector and $\Delta$ is a
$n \times m$ matrix built repeating $n$ times the row vector
$\delta = \bar{x}_n - \bar{x}_{n+1}$.
By definition $nQ^{}_{n+1} = Z_{n+1}^T Z^{}_{n+1}$ and, expanding the preceding
identity, we get
\begin{align}
n \, Q^{}_{n+1} \,
     = \, & \Sigma^{-1}_{n+1} \Sigma_n Z^T_n Z_n \Sigma_n \Sigma^{-1}_{n+1} +
            \nonumber \\
          & \Sigma^{-1}_{n+1} \Sigma_n (Z^T_n \Delta) \Sigma^{-1}_{n+1} +
            \nonumber \\
          & \Sigma^{-1}_{n+1} (\Delta^T Z_n) \Sigma_n \Sigma^{-1}_{n+1} +
            \nonumber \\
          & \Sigma^{-1}_{n+1} \Delta^T \Delta \Sigma^{-1}_{n+1} +
            \nonumber \\
          & z^T z
\end{align}
where $z=y\Sigma^{-1}_{n+1}$ and we used the fact that the $\Sigma$s are
diagonal.

Recalling that by hypothesis all the columns of the matrix $Z_n$ have zero mean
and that the columns of the matrix $\Delta$ have the same number, we see that
terms in parentheses are zero.
Thus
\begin{align}
n \, Q_{n+1} \, 
          = \, & \Sigma^{-1}_{n+1} \Sigma_n Q_n \Sigma_n \Sigma^{-1}_{n+1} +
                 \nonumber  \\
               & n \, \Sigma^{-1}_{n+1} \delta^T \delta \Sigma^{-1}_{n+1} +
                 z^T z
\end{align}
where $\delta^T \delta$, $z^T z$ and $Q_n$ are three $m \times m$ matrices.
We now see that we can compute $Q_{n+1}$ by making operations only on
$m \times m$ matrices and with the sole knowledge of $Q_n$ and $x_{n+1}$.

The computational advantage of this strategy is that we do not need to save
in the memory all the sampled data $X_{n+1}$ and moreover we do not need to
perform the explicit matrix product in eq. (\ref{eq:cov}), which would require
a great amount of memory and time for $n \approx 10^{5/6}$.
Consequently this algorithm can be fruitfully applied in situations where the
sensors number $m$ is small (e.g. of the order of tens) but the data stream is
expected to grow quickly.

The meaning of the normalization procedure depends on the process under
analysis and the meaning that is associated to the data within the current
study: both centering around the empirical mean and dividing by the empirical
variance can be avoided by respectively setting  $\Delta=0$ or $\Sigma=I$.

In practice, one keeps $n_{\rm start}$ observations and compute $Q$ as given by
eq. (\ref{eq:cov}) and the relative $C$ (and hence $P_{\rm start}$).
Then the updated $Q$s are used, step by step, to compute the $n^{th}$ values for
the evolving PCs in the standard way as $p_n = z_n C_n$.
In this way the last sample is equal for any $n$ to the one that would result
from a batch analysis until time step $n$.
Instead the whole sequence of the $p_n$ values with
$n_{\rm start} < n < n_{\rm final}$ would not coincide with those from
$P_{\rm final}$, since the $Q$s matrices change every time a sample is added,
and likewise for the $C$s matrices.
The most relevant implications of this fact will be considered in the next
subsection.

The library for the present implementation of the algorithm is available on the
Mathworks website under the name \textit{incremental PCA}.

\subsection{Continuity issues}
\label{sec:continuity}
\begin{figure*}[t!]
    \centering
    \begin{tabular}{c}
        \includegraphics[width=2.00\columnwidth]{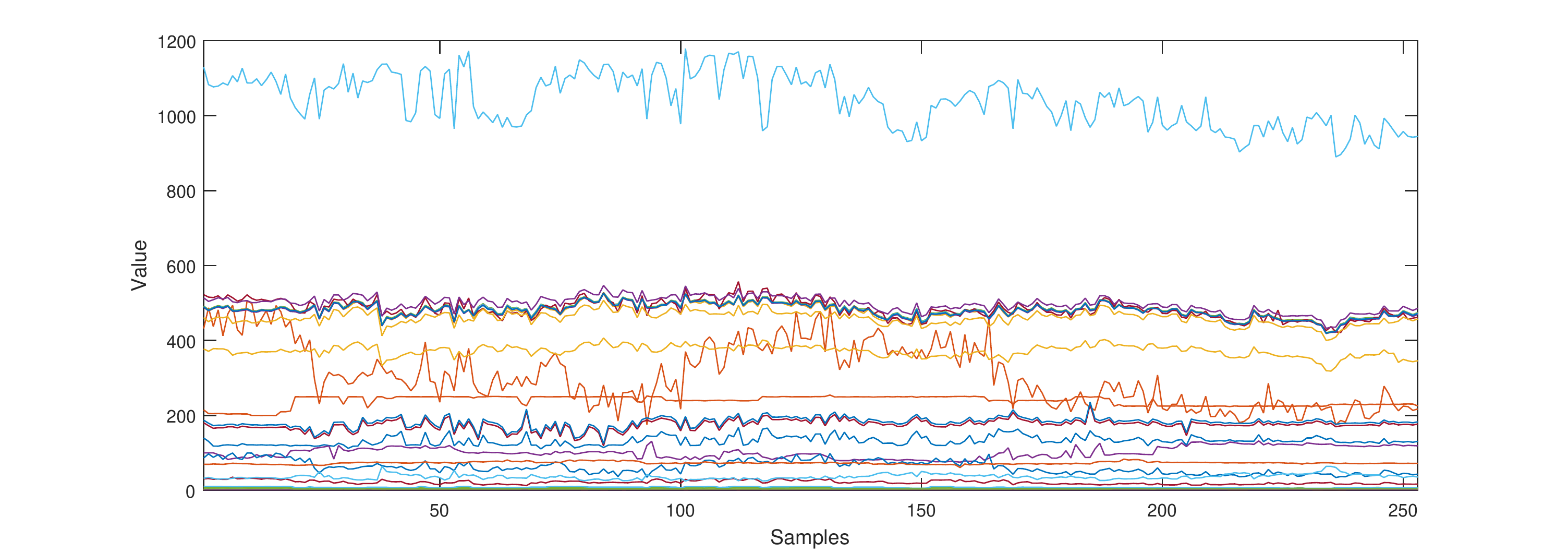} \\		
        \includegraphics[width=2.00\columnwidth]{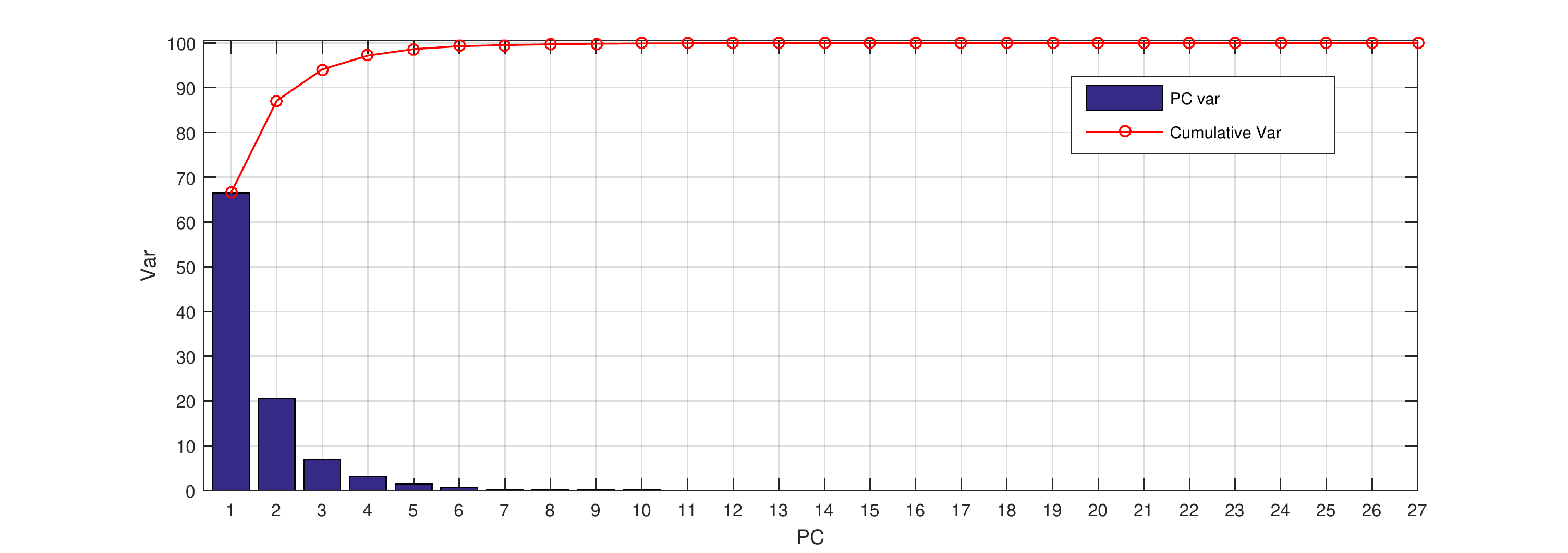}
    \end{tabular}
    \caption{Top, data-set used in the example.
             Bottom, variances associated to the PCs of the system.}
\label{fig:distillation}
\end{figure*}
We now consider the problem of the continuity for the PCs during the on-line
analysis.
In a batch analysis, one computes the PCs using all the data at the end of the
sampling, obtaining the matrix $C_{\rm final}$, and then, by applying this
transformation and its inverse, one can pass from the original data set to the
set PCs values.
Of course, since we are considering sampled data, we cannot speak of continuity in a strict sense. As previously stated, the temporal evolution of the data is not something relevant for the batch PCA. Regardless,  we may intuitively expect to use a sampling rate of at least two times the signal bandwidth (for the sampling theorem) usually even more, i.e. ten times. We hence expect a limited difference between two samples in proportion to the overall signal amplitude. For sampled data we can then define continuity in a intuitive sense as a condition where the difference between two consecutive samples is smaller than a given threshold. A discontinuity in the original samples may be reflected in the principal components depending on the transformation coefficients, in detail
\begin{equation}
	p_n-p_{n-1}=z_{n}C_{n}-z_{n-1}C_{n-1}
\end{equation}
that is equal to
\begin{equation}
\label{difference}
	p_n-p_{n-1}=(z_{n}-z_{n-1})C_{n}+z_{n-1}(C_{n}-C_{n-1})
\end{equation}
The first term would be the same for the batch procedure (in that case with constant $C$) and the second term shows how $p$ is changing due to the change in coefficients $C$. We can regard this term as the source of discontinuities due to the incremental algorithm.

To understand the problems that could arise, from the point of view of the
continuity of the PCs values, let us consider the on-line procedure more
closely.

We start with some the matrices $Q_{\rm start}$ and $C_{\rm start}$.
At a numerical level the eigenvalues are all different (since the machine
precision is at least of order $10^{-15}$), so that we have a set of formally
one-dimensional eigenspaces, from which the eigenvectors are taken.
Going on in the time sampling, we naturally create $m$ different time series of
eigenvectors.

\begin{figure*}[t!]
    \centering	
\includegraphics[width=2.00\columnwidth]{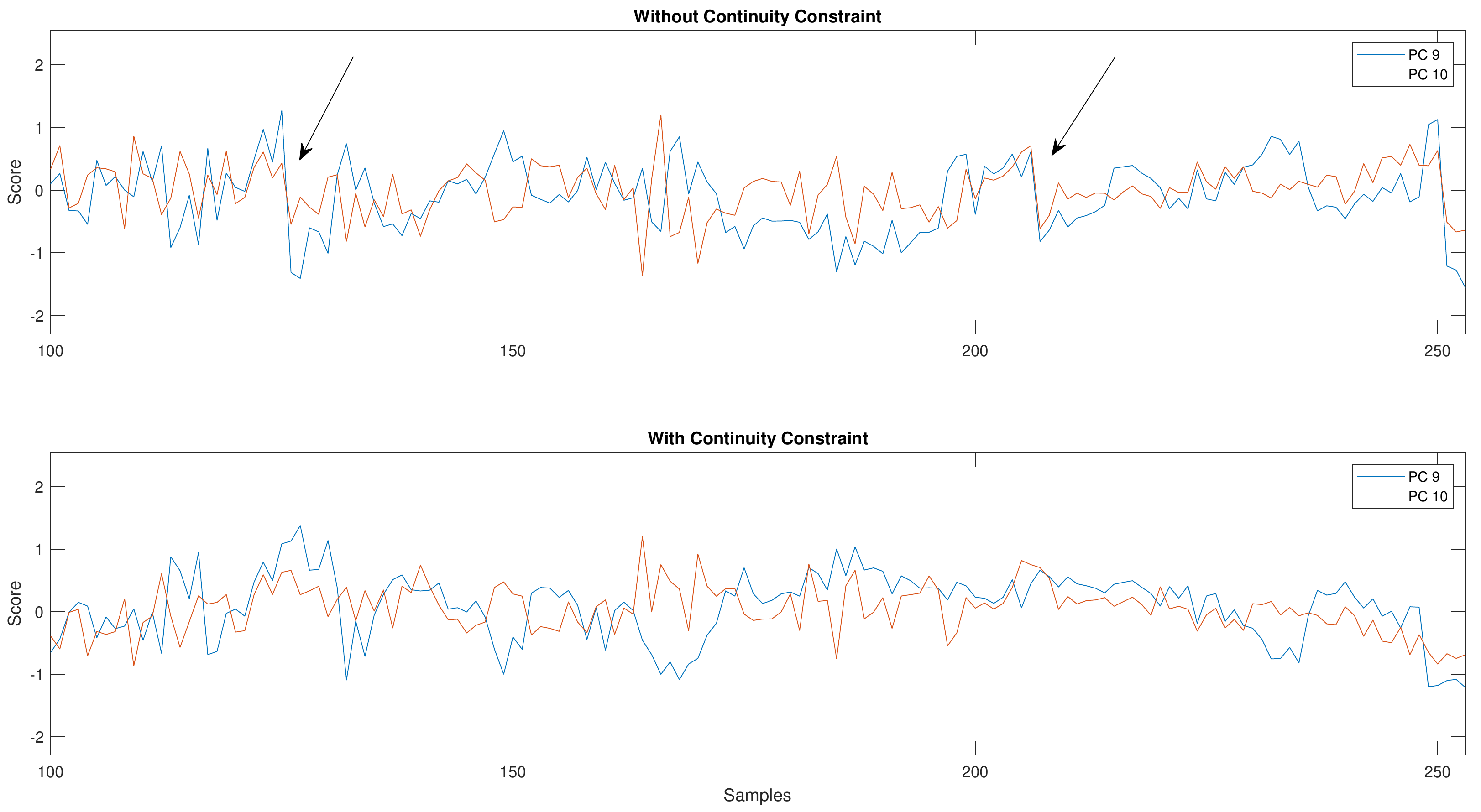}
    \caption{The figure shows the $9^{th}$ and the $10^{th}$ PCs computed without
             (top) and with (bottom) continuity constraints.
             Note the discontinuity addressed by the arrows.}
\label{fig:continuity}
\end{figure*}

We could expect that the difference of two subsequent eigenvectors of a given
series be slowly varying (in the sense of the standard euclidean norm), since
they come from different $C$s that are obtained from different $Qs$ which
differ only slightly (i.e. for the last $x_{n+1}$).
But this is not fully justified, since the PCs are not uniquely defined and in
some case two subsequent vectors of $p_n$ and $p_{n+1}$ can differ
considerably, as shown in the exaple in Figure \ref{fig:continuity}.
There are three ways in which one or more eigenvector series could exhibit a
discontinuity  (in the general sense discussed above).
\begin{itemize}
\item Consider the case of a given eigenvalue associated with two eigenspaces at two subsequent time steps, spanned by
      the vectors $c_n$ and $c_{n+1}$.
      They belong by hypothesis to two close ``lines'' but the algorithm can
      choose $c_{n+1}$ in the ``wrong'' direction.
      In this case, to preserve the on-line continuity as much as possible, we
      take the new PC to be $-c_{n+1}$, i.e. minus the eigenvector given by the
      diagonalization process at step $n+1$.
      The ``right'' orientation can be identified with simple considerations on
      the scalar product of $c_n$ with $c_{n+1}$.
      Recalling the considerations at the end of Section \ref{sec:PCA}, this
      substitution does not change the meaning of our analysis.
						
\item Consider the case where the differences of a group of $\nu$ contiguous
      eigenvalues are much smaller than the others: we can say that these
      eigenvalues correspond in fact to a degenerate eigenspace.
      In this case we can choose an infinite number of $\nu$ orthonormal
      vectors that can be legitimately considered our PCs, but the incremental
      algorithm can choose, at subsequent time steps, two basis which
      considerably differ.
      To overcome this problem, we must apply to the new transformation a
      change of basis in such a way not to modify the eigenspaces structure and
      to ``minimize'' the distance with the old basis. Although the case of a 
			proper degenerated space on real data is virtually impossible, as the difference 
			between two or more eigenvalues of $Q$ is approaching zero, the numerical values of the associated PCs can become
      discontinuous in a real time analysis. This by itself does not represent an error in an absolute sense in computing 
			$C_n$, as the	specific $C_n$ is the same as that which would be computed off-line.
			
\item In the two previous cases the discontinuity was due to an ambiguity of the diagonalization matrix. A third source of
			discontinuity can consist into the temporal evolution of the eigenvalues. Consider two one-dimensional eigenspaces associated, one with a variance that is increasing in time, the other with a variance that is decreasing: there will be a time step $\bar{n}$ the two eigenspaces are degenerate. This is called a ``level crossing'' and corresponds in the algorithm to an effective swap in the ``
			correct'' order of the eigenvectors. To restore continuity, the two components must me swapped.

\end{itemize}

\textcolor{black}{}
\section{Examples and Results}
\label{sec:results}
A publicly available data-set was used for this example: it consists of
snapshot measurements on $27$ variables from a distillation column, with a
sampling rate of one every three days, measured over $2.5$ years.
Sampling rate and time in general are not relevant \textit{per se} for PCA. Nevertheless, as we discussed the continuity issue it is interesting to see how the algorithm behaves on physical variables representing the continuous evolution of a physical system.

Variables represent temperatures, pressures, flows and other kind of measures
(the database is of industrial origin and the exact meaning of all the
variables is not specified).
Details are available on-line \cite{distillation}.

\begin{figure*}[t!]
    \centering
        \includegraphics[trim={5cm 2.8cm 5cm 0},clip,width=2.00\columnwidth]{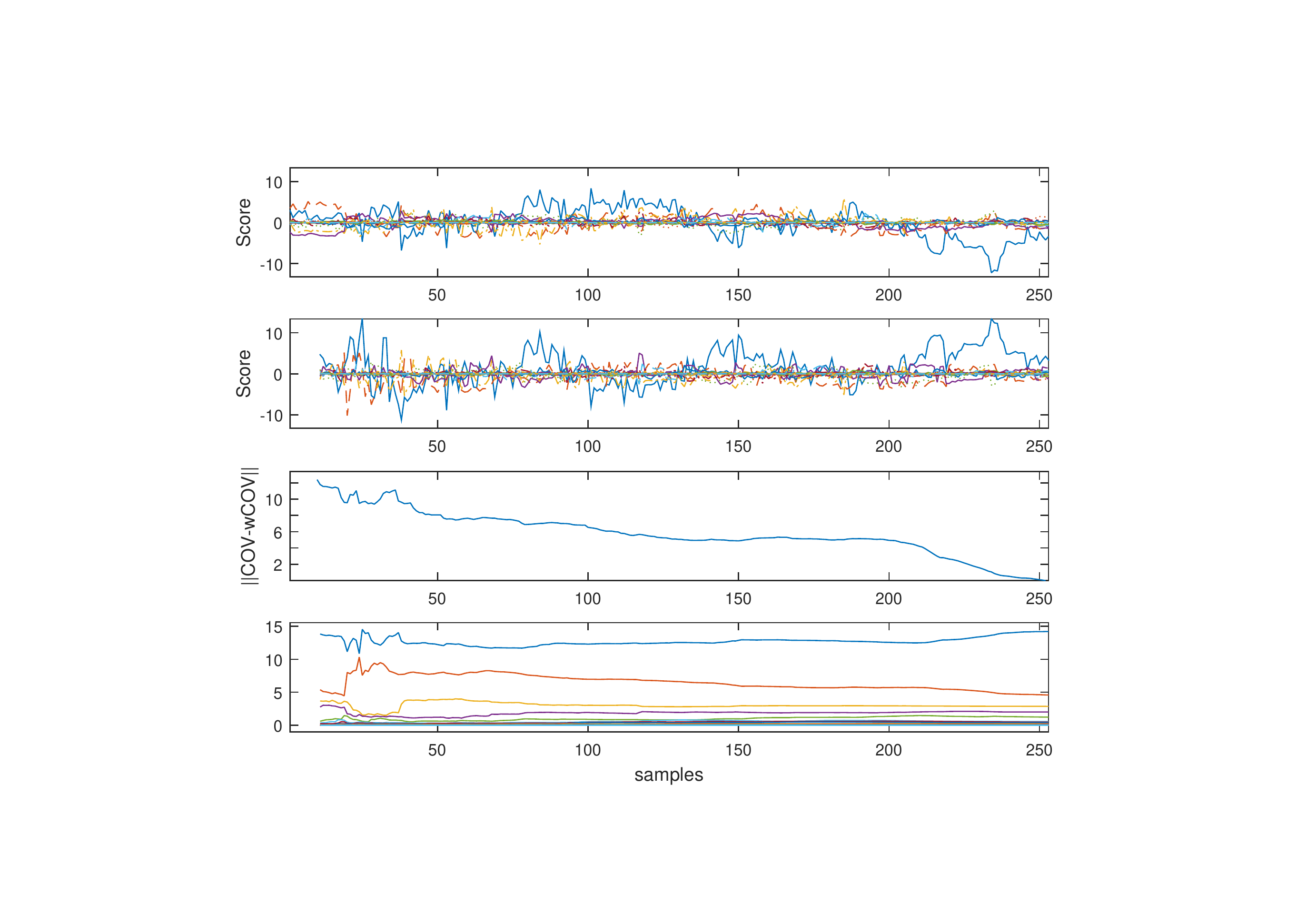}
    \caption{Application of the incremental algorithm to the sample data-set.
             The uppermost picture shows the 27 principal components computed
             with the batch algorithm, the second from top the incremental PCA
             computed without continuity check, the third picture from top
             represents the Frobenius norm of the difference between the
             covariance matrix computed through the incremental algorithm and a
             given sample and the one computed on the whole sample-set.
             The lowermost picture represents the variances associated to the
             PCs (eigenvalues of covariance matrix). \textcolor{black}{The covariance matrix and the variable values are the same for the batch algorithm and the on-line implementation when they are provided with the same samples. The differences in the pictures are due to the fact that same transformation computed with the batch algorithm is applied to the whole set, while the one computed online changes with every sample.}}
\label{fig:resultDistillation}
\end{figure*}

This kind of data set includes variables that are strongly correlated amongst
each other, variables with a large variance and variables almost constant during
a time of several samples.
In Figure \ref{fig:distillation} we display the time evolution of the variables
and the standard batch PCA.
In Figure \ref{fig:resultDistillation} the evolution of the covariance matrix
$Q$ and the incremental PCs are shown.
Notice that the values $p_n$ are obviously not equal to the ones computed with
the batch method until the last sample.
The matrix $Q$ almost constantly converges to the covariance matrix computed
with the batch method.
Note that at the beginning the Frobenius norm of the difference between the
two matrices sometimes grows with the addition of some samples, the number of
samples needed for $Q$ to settle to the final value depends on the regularity
of the data and the variations in $Q$ may represent an interesting description
of the analyzed process.
This is expected for the estimator of the covariance matrix until $m \gtrsim n$.
While the sample covariance matrix is an unbiased estimator for
$n \rightarrow \infty$, it is known to converge  inefficiently
\cite{smith2005covariance}.

\begin{figure}[htb]
	\centering
		\includegraphics[width=1.00\columnwidth]{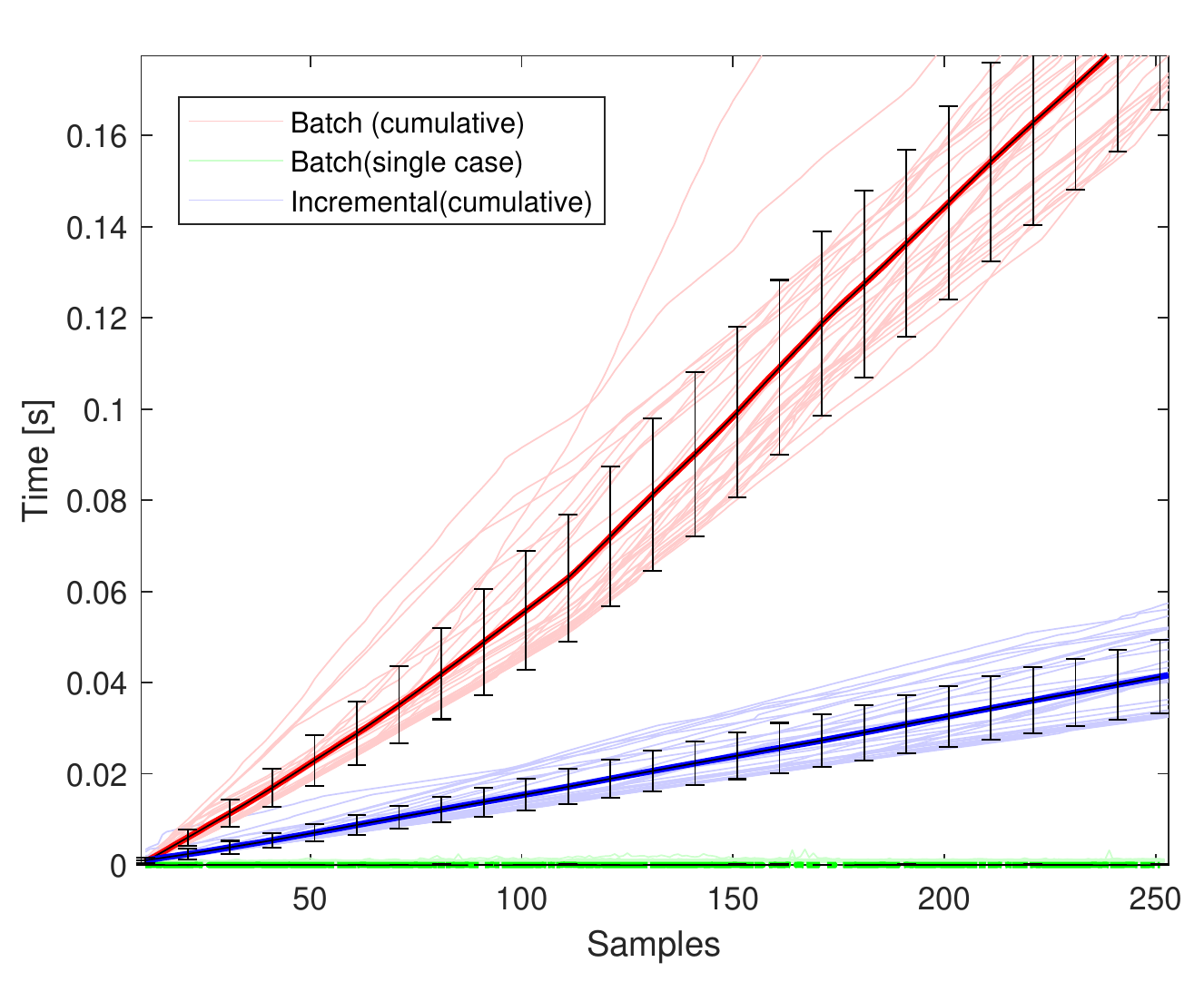}
	\caption{Time required to execute the incremental PCA and the batch implementation as function of the number of samples. For the batch algorithm both the time required to compute the PCA on the given number of samples (single case) and the cumulative time required to perform the PCA with each additional sample (cumulative) are shown. The computational time is measured empirically and can be affected by small fluctuations due to the activity of the operative system: in order to take this in account the average times (darker lines) and their standard deviations (error bars) are computed on 33 trials. The batch implementation is more efficient than the incremental one when the PCA is computed on the whole dataset, while the incremental implementation is more efficient when samples are added incrementally.}
	\label{fig:PerformanceTcrop}
\end{figure}

In order to quantify the efficiency of the algorithm the computational of the proposed incremental solution has been compared with the batch implementation provided by the Matlab built-in function \textit{PCA} on a Intel Core i7-7700HQ CPU running at 2.81 GHz, with windows 10 operative system. The results are shown in Figure \ref{fig:PerformanceTcrop}. The time required to execute the incremental algorithm grows linearly with the number of samples while the batch presents an increase of the execution time associated with the size of the dataset. As reasonably expected, the batch implementation is more efficient than the incremental one when the PCA is computed on the whole dataset, while the incremental implementation is more efficient when samples are added incrementally. 

\begin{figure}[htb]
    \centering
 \includegraphics[width=1.00\columnwidth]{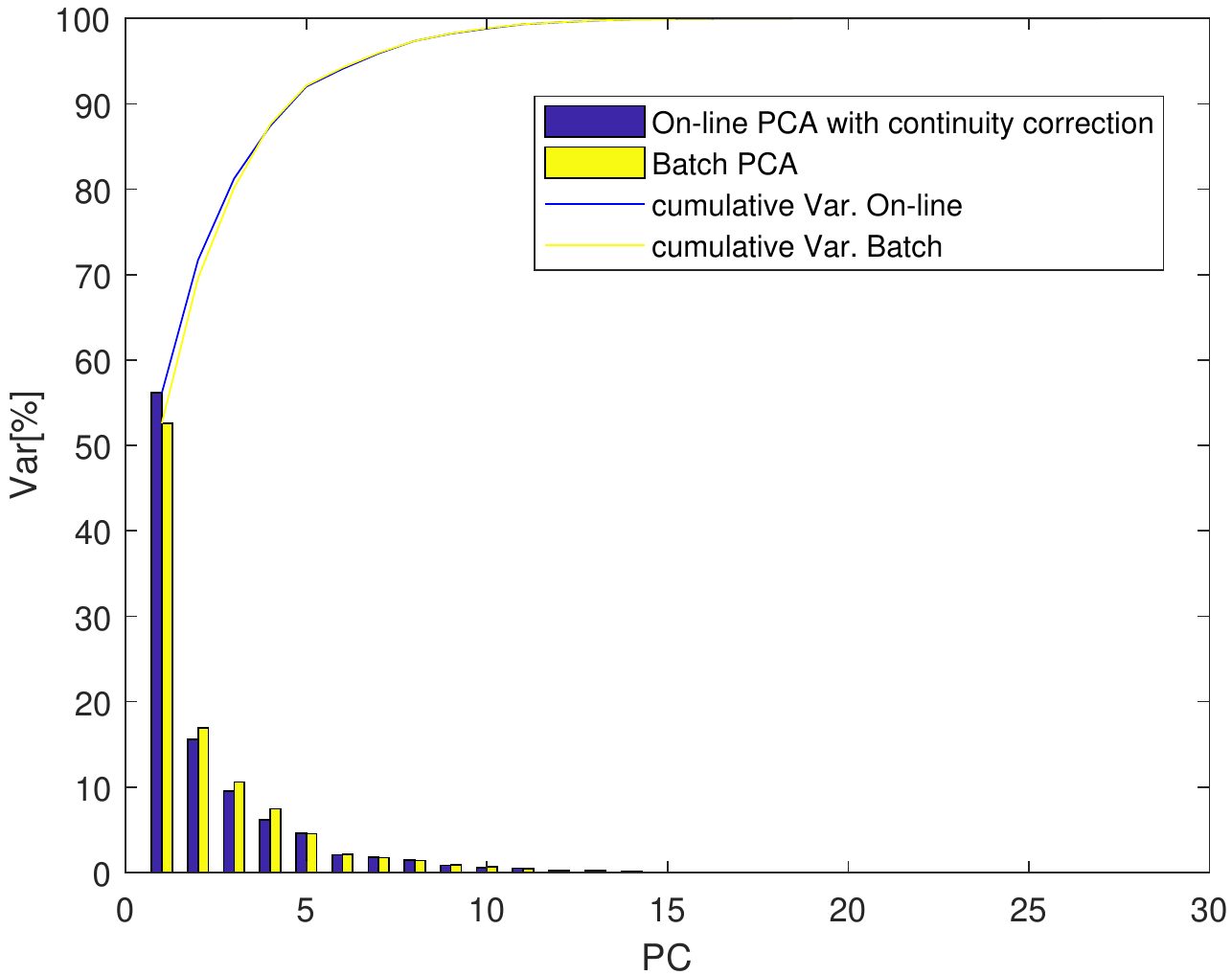}
    \caption{Variances and cumulative variance of the
             PCs computed with the on-line algorithm \textcolor{black}{including the continuity correction (blue) and the ones computed with the batch algorithm (yellow)}.The order of the on-line computed PCs is the one produced by the transformation.\textcolor{black}{. The difference between the two set of PCs' variances are due to the continuity correction and the fact that the variance of the on-line series is computed on the whole set of data.}}
\label{fig:postPC}
\end{figure}

In Figure \ref{fig:postPC} the variance of the whole incremental PCA is shown.
Comparing it with Figure \ref{fig:distillation} (bottom), it is evident that
the incremental PCs that are not linearly independent over the whole sampling
time have a slightly different distribution of the variance compared to the PCs
computed with the batch algorithm.
Nevertheless they are a good approximation in that they are still ordered by
variance and most of the variance is in the first components
(i.e. more than $90 \%$ is in the first 5 PCs).

\section{Discussion and Conclusions}
\label{sec:conclusion}
The continuity issues arise for principal components with similar
variances.
When working with real data this issue often affects the components with
smaller variance which are usually dropped and hence it can be reasonable to
execute the algorithm without taking measures to preserve the continuity.

Nevertheless it should be noticed that, in some process analysis, the
components with a smaller variance identify the \textit{stable} part of the
analyzed data, and hence the one identifying the process, e.g. the controlled
variables in a human movement \cite{lippi2011uncontrolled} or the  response of
a dynamic system known through the input-output samples \cite{huang2001process}.

In Figure \ref{fig:continuity} the effects of discontinuities are shown: two discontinuities present into the values of one of the principal components are fixed according to Section \ref{sec:continuity}.
In case the continuity is imposed the phenomenon is limited, but this comes at
the price of modifying the exact computation of the eigenvectors for $Q$ at a
given step, in case of a degenerate eigenspace.
Anyway the error introduced on the $Q$ eigenvectors depends on the threshold
used to establish that two slightly different eigenvalues are degenerate and so
we can still consider the transformation to be ``exact'', but not at machine precision.
In the reported example, the two big discontinuities highlighted by arrows disappear when the continuity is imposed. Notice that the two PCs have different values in the corrected version also before the two big discontinuities because of previous corrections on $Q$.
The choice as to whether or not the continuity is imposed  depends on the application, on the
data-set and on the meaning associated with the analysis.
\subsection{Software}
The \textsc{Matlab} software implementing the function and the examples shown in the figures is available at the URL: 
\href{https://it.mathworks.com/matlabcentral/fileexchange/69844-incremental-principal-component-analysis}{https://it.mathworks.com/matlabcentral/fileexchange/ 69844-incremental-principal-component-analysis}

\section*{Acknowledgments}
The authors thank Prof. Thomas Mergner for the support to this work. \\
The financial support from the European project H$_{2}$R \\
(http://www.h2rproject.eu/) is appreciated. \\
We gratefully acknowledge financial support for the project MTI-engAge (16SV7109) by BMBF \\
G.C. has been supported by I.N.F.N. \\
\balance
\bibliographystyle{apalike}
{
\bibliography{example}}

\end{document}